%% file: main.tex
\documentclass[11pt,a4paper]{amsart}

\usepackage[margin=1in]{geometry}
\usepackage{amsmath,amssymb,amsthm}
\IfFileExists{mathtools.sty}{\usepackage{mathtools}}{\PackageWarning{main}{mathtools not installed; continuing with amsmath fallback}}
\usepackage[authoryear,round]{natbib}
\usepackage{booktabs,graphicx,array,tabularx}
\IfFileExists{caption.sty}{\usepackage[font=small,labelfont=bf,skip=8pt]{caption}}{}
\usepackage{xcolor}
\usepackage{microtype}
\usepackage[hidelinks]{hyperref}
\graphicspath{{figures/}}
\numberwithin{equation}{section}

\newtheorem{theorem}{Theorem}[section]
\newtheorem{lemma}[theorem]{Lemma}
\newtheorem{proposition}[theorem]{Proposition}
\newtheorem{corollary}[theorem]{Corollary}
\theoremstyle{definition}

\theoremstyle{remark}
\newtheorem{remark}[theorem]{Remark}

\DeclareMathOperator{\tr}{tr}

\newcommand{\R}{\mathbb{R}}

\newcommand{\cC}{\mathcal{C}}
\newcommand{\cD}{\mathcal{D}}

\newcommand{\one}{\mathbf{1}}
\newcommand{\T}{\mathsf{T}}
\newcommand{\op}{o_{P}}

\newcommand{\astar}{a^{\star}}
\newcommand{\bstar}{b^{\star}}
\newcommand{\sstar}{s^{\star}}

\title[Outliers and the Hard-Margin Linear SVM in HDLSS]{Outliers and the Hard-Margin Linear SVM\\in High-Dimension, Low-Sample-Size Settings}
\author{Yugo Nakayama}
\date{}

\makeatletter
\let\original@setauthors\@setauthors
\def\@setauthors{%
  \original@setauthors
  \vspace{-0.1em}\begin{center}\small\mdseries Independent researcher\end{center}%
}
\def\@setaddresses{}
\makeatother

\begin{document}
\begin{abstract}
We study how outliers affect the hard-margin linear support vector machine when the dimension grows while the sample sizes remain fixed. For one-sided contamination, the centered Gram matrix converges to a constant inlier block plus an outlier-diagonal component, and the limiting dual problem has a closed-form solution. All outliers act through one effective distance. This representation yields explicit thresholds for loss of inlier support, margin shrinkage, intercept displacement, and asymptotic misclassification. We then treat contamination in both classes. The limiting intercept is the multiplier of the class-balance constraint, and the two classes pull it in opposite directions. Consequently, at most one class can collapse or be asymptotically misclassified, whereas symmetric contamination leaves the limiting discriminant function unchanged even though the margin shrinks. Seven numerical experiments verify the dual allocation, phase transitions, and discriminant limits.
\end{abstract}

\maketitle

\input{sections/01_intro}
\input{sections/02_setup}
\input{sections/03_geometry}
\input{sections/04_limiting_dual}
\input{sections/05_one_sided}
\input{sections/06_discriminant}
\input{sections/07_two_sided}
\input{sections/08_simulations}
\input{sections/09_discussion}

\section*{Acknowledgments}
Parts of the drafting and technical editing of this manuscript were assisted by large language models, including Anthropic's Claude Opus 5.5. The author takes full responsibility for the correctness of all mathematical statements, proofs, and numerical results.

\bibliographystyle{plainnat}
\bibliography{references}
\end{document}

%% file: sections/01_intro.tex
\section{Introduction}

Genetic microarrays, medical imaging, and robotics routinely produce high-dimension, low-sample-size (HDLSS) data, in which the dimension far exceeds the sample size. \citet{hall2005} showed that such data acquire a rigid geometry as the dimension diverges, and \citet{ahn2007} proved that this representation holds under mild conditions; see also \citet{jung2009} and the survey of \citet{aoshima2018}. Two lines of work build on this geometry for classification. One develops linear discriminant rules for many variables \citep{bickel2004,fan2008} and distance-based or geometric classifiers \citep{hall2008,chan2009,aoshima2011,aoshima2014,aoshima2015,aoshima2019}. The other studies margin-based methods, including distance-weighted discrimination \citep{marron2007,qiao2010} and the maximal data-piling direction \citep{ahn2010}.

Within the second line, \citet{nakayama2017} derived the asymptotic behavior of the hard-margin linear support vector machine (SVM) \citep{cortes1995,vapnik2000}. They showed that the SVM is consistent under suitable conditions but can be heavily biased when the classes differ in covariance trace or sample size, and proposed a bias-corrected SVM. \citet{nakayama2020} extended that analysis to the Gaussian kernel.

These results assume that every training point comes from one of the two populations. Real training data can contain mislabeled samples or measurements taken under abnormal conditions \citep{frenay2014}. Robust statistics has a long history \citep{huber1964,hampel1986,rousseeuw1987}, and SVM robustness has been studied through influence functions \citep{christmann2004,steinwart2008}, robust losses \citep{wu2007}, robust optimization \citep{xu2009}, and poisoning attacks \citep{biggio2012}. Those analyses either fix the dimension or use conditions that do not reflect HDLSS geometry.

Suppose one class contains a point whose mean $\boldsymbol\mu_0$ lies far from its nominal mean $\boldsymbol\mu_2$. Can the SVM learn the contrast $\boldsymbol\mu_2-\boldsymbol\mu_0$ instead of the population contrast $\boldsymbol\mu_1-\boldsymbol\mu_2$? Because the hard-margin SVM depends only on support vectors, a single outlier may change the classifier completely. A second question is what happens when both classes are contaminated.

Our contributions are as follows.
\begin{itemize}
  \item We derive the limit of the centered Gram matrix in the presence of outliers (Lemma~\ref{lem:geometry}).
  \item For one-sided contamination, we solve the limiting dual problem in closed form and show that the outliers enter only through one effective distance (Lemma~\ref{lem:reduction} and Theorem~\ref{thm:one-sided}).
  \item We identify a sharp threshold below which the contaminated class loses all inlier support and obtain the corresponding critical number of outliers (Corollary~\ref{cor:allocation}).
  \item We prove convergence of the actual dual value, dual allocation, intercept, and discriminant function, and derive limiting misclassification rates (Theorems~\ref{thm:transfer-value}--\ref{thm:discriminant} and Corollary~\ref{cor:misclass-one}).
  \item For two-sided contamination, we solve the KKT system, identify its multiplier with the limiting intercept, and show that at most one class can fail. Symmetric contamination cancels the intercept shift (Theorem~\ref{thm:two-sided} and Corollaries~\ref{cor:breakdown-two}--\ref{cor:symmetric}).
  \item Seven simulation studies check the finite-dimensional agreement with these limits (Section~\ref{sec:simulations}).
\end{itemize}

All algebraic steps use only linear algebra, probability, and convex optimization \citep{boyd2004}.

%% file: sections/02_setup.tex
\section{Setup}
\subsection{Populations and outliers}

Let $\Pi_1$ and $\Pi_2$ be independent $d$-variate populations. Population $\Pi_i$ has mean $\boldsymbol\mu_i$ and covariance matrix $\boldsymbol\Sigma_i$. From $\Pi_i$ we observe independent vectors $\boldsymbol x_{i1},\ldots,\boldsymbol x_{in_i}$. Until Section~\ref{sec:two-sided}, the training data labeled as class 2 additionally contain $m$ independent outliers $\boldsymbol y_1,\ldots,\boldsymbol y_m$, where $\boldsymbol y_k$ has mean $\boldsymbol\mu_{0k}$ and covariance $\boldsymbol\Sigma_{0k}$. We hold $n_1,n_2,m$ fixed and let $d\to\infty$.

Let $I_1,I_2$, and $O$ index the class-1 inliers, class-2 inliers, and outliers. The full training set is $\boldsymbol x_1,\ldots,\boldsymbol x_N$, where $N=n_1+n_2+m$, with labels $t_j=-1$ for $j\in I_1$ and $t_j=+1$ for $j\in I_2\cup O$. Define
\begin{equation}
\boldsymbol\mu=\boldsymbol\mu_1-\boldsymbol\mu_2,\quad
\Delta=\lVert\boldsymbol\mu\rVert^2,\quad
\boldsymbol\mu_*=\frac{\boldsymbol\mu_1+\boldsymbol\mu_2}{2},\quad
\tau_i=\frac{\tr(\boldsymbol\Sigma_i)}{n_i},\quad
\Delta_*=\Delta+\tau_1+\tau_2.
\end{equation}
For outlier $k$, let
\begin{equation}
\boldsymbol\nu_k=\boldsymbol\mu_{0k}-\boldsymbol\mu_*,\qquad
D_k=\lVert\boldsymbol\nu_k\rVert^2+\tr(\boldsymbol\Sigma_{0k}),\qquad
\kappa=\left(\sum_{k=1}^{m}\frac{1}{D_k}\right)^{-1}.
\end{equation}
We call $D_k$ the effective squared distance of outlier $k$ and $\kappa$ the effective outlier distance. If $D_1=\cdots=D_m=D$, then $\kappa=D/m$; thus $m$ equidistant outliers act like one outlier that is $m$ times closer.

\subsection{Hard-margin linear SVM}
The hard-margin linear SVM minimizes $\lVert\boldsymbol w\rVert^2/2$ subject to $t_j(\boldsymbol w^\T\boldsymbol x_j+b)\ge1$. Its dual is
\begin{equation}\label{eq:dual}
\max_{\boldsymbol\alpha}
L(\boldsymbol\alpha)=\sum_{j=1}^{N}\alpha_j-\frac12\sum_{j,k=1}^{N}\alpha_j\alpha_k t_jt_k\boldsymbol x_j^\T\boldsymbol x_k,
\quad
\alpha_j\ge0,
\quad
\sum_{j=1}^{N}\alpha_jt_j=0.
\end{equation}
Writing $\boldsymbol t=(t_1,\ldots,t_N)^\T$, let
\[
\cC=\{\boldsymbol\alpha\in\R^N:\boldsymbol\alpha\ge0,\ \boldsymbol t^\T\boldsymbol\alpha=0\}.
\]
If $\widehat{\boldsymbol\alpha}$ maximizes \eqref{eq:dual}, then
$\widehat{\boldsymbol w}=\sum_j\widehat\alpha_jt_j\boldsymbol x_j$ and
$\widehat y(\boldsymbol x)=\widehat{\boldsymbol w}^{\T}\boldsymbol x+\widehat b$.
The KKT conditions give $t_j\widehat y(\boldsymbol x_j)=1$ whenever $\widehat\alpha_j>0$.

\subsection{Assumptions}
Let $\boldsymbol\varepsilon_j=\boldsymbol x_j-\boldsymbol\mu_i$ for $j\in I_i$ and $\boldsymbol\eta_k=\boldsymbol y_k-\boldsymbol\mu_{0k}$. As $d\to\infty$, assume:
\begin{description}
\item[(A1)] For $i,l\in\{1,2\}$,
\[
{\tr(\boldsymbol\Sigma_i\boldsymbol\Sigma_l)\over\Delta_*^2}\to0,
\qquad
{\boldsymbol\mu^\T\boldsymbol\Sigma_i\boldsymbol\mu\over\Delta_*^2}\to0,
\qquad
{\operatorname{Var}(\lVert\boldsymbol\varepsilon\rVert^2)\over\Delta_*^2}\to0
\quad(\boldsymbol\varepsilon\sim(0,\boldsymbol\Sigma_i)).
\]
\item[(A2)] For $k\ne l$ and $i=1,2$,
\[
{\lvert\boldsymbol\mu^\T\boldsymbol\nu_k\rvert+\lvert\boldsymbol\nu_k^\T\boldsymbol\nu_l\rvert\over\Delta_*}\to0,
\]
\[
{\boldsymbol\mu^\T\boldsymbol\Sigma_{0k}\boldsymbol\mu+
\boldsymbol\nu_k^\T\boldsymbol\Sigma_i\boldsymbol\nu_k+
\tr(\boldsymbol\Sigma_i\boldsymbol\Sigma_{0k})+
\boldsymbol\nu_k^\T\boldsymbol\Sigma_{0l}\boldsymbol\nu_k+
\tr(\boldsymbol\Sigma_{0k}\boldsymbol\Sigma_{0l})
\over\Delta_*^2}\to0,
\]
\[
{\boldsymbol\nu_k^\T\boldsymbol\Sigma_{0k}\boldsymbol\nu_k+
\operatorname{Var}(\lVert\boldsymbol\eta_k\rVert^2)
\over D_k^2}\to0.
\]
\item[(A3)] $\limsup_{d\to\infty}\tau_2/(\Delta+\tau_1)<\infty$.
\end{description}
Condition (A1) is standard in HDLSS analysis. For example, it holds for identity covariance and bounded fourth moments. Condition (A2) requires the outlier deviations to be asymptotically orthogonal to the population contrast and to each other, which is typical for unrelated directions in high dimensions. Section~\ref{sec:two-sided} introduces (A4).

\subsection{Notation guide}
Table~\ref{tab:notation} collects the symbols used repeatedly below.
\begin{table}[htbp]
\centering
\caption{Principal notation.}
\label{tab:notation}
\begin{tabularx}{\textwidth}{@{}>{$}l<{$}X@{}}
\toprule
\text{Symbol} & \text{Meaning}\\
\midrule
n_i,\ I_i & Inlier sample size and index set for class $i$.\\
O,\ O_i,\ m_i & Outlier index set; in Section~\ref{sec:two-sided}, $O_i$ and $m_i=|O_i|$ refer to class $i$.\\
\Delta,\ \Delta_* & Squared mean separation and its noise-adjusted version.\\
\tau_i & Per-sample covariance trace, $\tr(\boldsymbol\Sigma_i)/n_i$.\\
D_k,\ \kappa,\ \kappa_i & Effective squared distance of one outlier; harmonic aggregate for one or for class $i$.\\
p & Shorthand $p=\Delta/4$.\\
a,b,s & Total dual mass on class-1 inliers, class-2 inliers, and class-2 outliers.\\
s_i & Total dual mass on outliers labeled as class $i$.\\
E & $\Delta(\tau_1+\tau_2)/4+\tau_1\tau_2$.\\
H & $\kappa_1\kappa_2\Delta_*+(\kappa_1+\kappa_2)E$ in the two-sided problem.\\
\rho_i & Parallel effective cost $\{(p+\tau_i)^{-1}+\kappa_i^{-1}\}^{-1}$.\\
\beta_i(x) & Two-sided collapse boundary $px/(x+p+\tau_{3-i})$.\\
(a^\circ,b^\circ) & Unconstrained stationary point of the reduced one-sided objective.\\
(a^\star,b^\star) & Constrained maximizer over $\cD=\{(a,b):a\ge b\ge0\}$.\\
\widetilde{\boldsymbol G},\ \boldsymbol R & Deterministic Gram limit and remainder $\boldsymbol G-\widetilde{\boldsymbol G}$.\\
\bottomrule
\end{tabularx}
\end{table}

%% file: sections/03_geometry.tex
\section{Translation invariance and geometric representation}

\begin{lemma}[Translation invariance]\label{lem:translation}
For any $\boldsymbol c\in\R^d$ and $\boldsymbol\alpha$ satisfying $\boldsymbol t^\T\boldsymbol\alpha=0$,
\[
\sum_{j,k}\alpha_j\alpha_k t_jt_k\boldsymbol x_j^\T\boldsymbol x_k
=
\sum_{j,k}\alpha_j\alpha_k t_jt_k(\boldsymbol x_j-\boldsymbol c)^\T(\boldsymbol x_k-\boldsymbol c).
\]
\end{lemma}
\begin{proof}
Bilinearity gives
\[
\sum_{j,k}\alpha_j\alpha_kt_jt_k\boldsymbol x_j^\T\boldsymbol x_k
=\left\lVert\sum_j\alpha_jt_j\boldsymbol x_j\right\rVert^2.
\]
Moreover,
\[
\sum_j\alpha_jt_j\boldsymbol x_j
=\sum_j\alpha_jt_j(\boldsymbol x_j-\boldsymbol c)
+\left(\sum_j\alpha_jt_j\right)\boldsymbol c
=\sum_j\alpha_jt_j(\boldsymbol x_j-\boldsymbol c).
\]
\end{proof}

Take $\boldsymbol c=\boldsymbol\mu_*$ and define the signed centered Gram matrix $\boldsymbol G=(G_{jk})$ by
\[
G_{jk}=t_jt_k(\boldsymbol x_j-\boldsymbol\mu_*)^\T(\boldsymbol x_k-\boldsymbol\mu_*).
\]
Then
\begin{equation}\label{eq:centered-dual}
L(\boldsymbol\alpha)=\one^\T\boldsymbol\alpha-\frac12\boldsymbol\alpha^\T\boldsymbol G\boldsymbol\alpha,
\qquad \boldsymbol\alpha\in\cC.
\end{equation}
Its deterministic leading term is
\begin{equation}\label{eq:Gtilde}
\widetilde G_{jk}=
\begin{cases}
\Delta/4+\delta_{jk}\tr(\boldsymbol\Sigma_i),&j,k\in I_1\cup I_2,\ j\in I_i,\\
D_k\delta_{jk},&j,k\in O,\\
0,&\text{otherwise},
\end{cases}
\end{equation}
where $\delta_{jk}$ is the Kronecker delta.

\begin{lemma}[Geometric representation with outliers]\label{lem:geometry}
Assume (A1)--(A2), and let $\boldsymbol R=\boldsymbol G-\widetilde{\boldsymbol G}$. Then
\[
\epsilon_1:=\max_{(j,k)\notin O\times O}{|R_{jk}|\over\Delta_*}=\op(1),
\qquad
\epsilon_2:=\max_{k\in O}{|R_{kk}|\over D_k}=\op(1).
\]
\end{lemma}
\begin{proof}
Put $\boldsymbol u=-\boldsymbol\mu/2$. For every inlier,
$t_j(\boldsymbol x_j-\boldsymbol\mu_*)=\boldsymbol u+t_j\boldsymbol\varepsilon_j$,
whereas for $k\in O$,
$t_k(\boldsymbol y_k-\boldsymbol\mu_*)=\boldsymbol\nu_k+\boldsymbol\eta_k$.
Also $\lVert\boldsymbol u\rVert^2=\Delta/4$.

For distinct inliers $j\in I_i$ and $k\in I_l$,
\[
R_{jk}=t_j\boldsymbol u^\T\boldsymbol\varepsilon_j
+t_k\boldsymbol u^\T\boldsymbol\varepsilon_k
+t_jt_k\boldsymbol\varepsilon_j^\T\boldsymbol\varepsilon_k.
\]
The three terms have mean zero. Their second moments are controlled by
$\boldsymbol\mu^\T\boldsymbol\Sigma_i\boldsymbol\mu/4$,
$\boldsymbol\mu^\T\boldsymbol\Sigma_l\boldsymbol\mu/4$, and
$\tr(\boldsymbol\Sigma_i\boldsymbol\Sigma_l)$; Chebyshev's inequality and (A1) therefore give $R_{jk}/\Delta_*=\op(1)$.
For an inlier diagonal,
\[
R_{jj}=2t_j\boldsymbol u^\T\boldsymbol\varepsilon_j+
\{\lVert\boldsymbol\varepsilon_j\rVert^2-\tr(\boldsymbol\Sigma_i)\},
\]
and (A1) gives the same conclusion.

For $j\in I_i$ and $k\in O$,
\[
R_{jk}=G_{jk}=-{\boldsymbol\mu^\T\boldsymbol\nu_k\over2}
-{\boldsymbol\mu^\T\boldsymbol\eta_k\over2}
+t_j\boldsymbol\varepsilon_j^\T\boldsymbol\nu_k
+t_j\boldsymbol\varepsilon_j^\T\boldsymbol\eta_k.
\]
Assumption (A2) makes the deterministic term negligible relative to $\Delta_*$ and bounds the second moments of the other terms. For $k\ne l$ in $O$,
$R_{kl}=\boldsymbol\nu_k^\T\boldsymbol\nu_l+\boldsymbol\nu_k^\T\boldsymbol\eta_l+\boldsymbol\nu_l^\T\boldsymbol\eta_k+\boldsymbol\eta_k^\T\boldsymbol\eta_l$, which is again $\op(\Delta_*)$ by (A2). Finally,
\[
R_{kk}=2\boldsymbol\nu_k^\T\boldsymbol\eta_k+
\{\lVert\boldsymbol\eta_k\rVert^2-\tr(\boldsymbol\Sigma_{0k})\}=\op(D_k).
\]
Because $N$ is fixed, a union bound over the finitely many entries proves both maxima.
\end{proof}

%% file: sections/04_limiting_dual.tex
\section{The limiting dual problem}
Equation~\eqref{eq:centered-dual} and Lemma~\ref{lem:geometry} motivate
\begin{equation}\label{eq:Ltilde}
\widetilde L(\boldsymbol\alpha)=\one^\T\boldsymbol\alpha-\frac12\boldsymbol\alpha^\T\widetilde{\boldsymbol G}\boldsymbol\alpha,
\qquad \boldsymbol\alpha\in\cC.
\end{equation}
For $\boldsymbol\alpha\in\cC$, define
\[
a=\sum_{j\in I_1}\alpha_j,\qquad
b=\sum_{j\in I_2}\alpha_j,\qquad
s=\sum_{k\in O}\alpha_k.
\]
The equality constraint is $-a+b+s=0$, so $s=a-b$. Thus $(a,b)$ ranges over
$\cD=\{(a,b):a\ge b\ge0\}$ and $\one^\T\boldsymbol\alpha=2a$.

\subsection{A weighted Cauchy--Schwarz inequality}
\begin{lemma}\label{lem:weighted-cs}
Let $w_1,\ldots,w_r>0$. For real $z_1,\ldots,z_r$ with $\sum_jz_j=S$,
\[
\sum_{j=1}^{r}w_jz_j^2\ge {S^2\over\sum_jw_j^{-1}},
\]
with equality if and only if $z_j=Sw_j^{-1}/\sum_lw_l^{-1}$.
\end{lemma}
\begin{proof}
Apply Cauchy--Schwarz to $\sqrt{w_j}z_j$ and $1/\sqrt{w_j}$:
$S^2\le(\sum_jw_jz_j^2)(\sum_jw_j^{-1})$. The equality condition gives the stated allocation.
\end{proof}

\subsection{Reduction to two variables}
Put $p=\Delta/4$ and define
\begin{equation}\label{eq:QF}
Q(a,b)=p(a+b)^2+\tau_1a^2+\tau_2b^2+\kappa(a-b)^2,
\qquad
F(a,b)=2a-\frac12Q(a,b).
\end{equation}
\begin{lemma}[Reduction]\label{lem:reduction}
For every $\boldsymbol\alpha\in\cC$ with classwise sums $(a,b)$,
$\boldsymbol\alpha^\T\widetilde{\boldsymbol G}\boldsymbol\alpha\ge Q(a,b)$. Moreover,
\[
\max_{\boldsymbol\alpha\in\cC}\widetilde L(\boldsymbol\alpha)
=
\max_{(a,b)\in\cD}F(a,b),
\]
and, for each $(a,b)\in\cD$, equality is attained by
\begin{equation}\label{eq:allocation}
\alpha_j={a\over n_1}\ (j\in I_1),\qquad
\alpha_j={b\over n_2}\ (j\in I_2),\qquad
\alpha_k=(a-b){\kappa\over D_k}\ (k\in O).
\end{equation}
\end{lemma}
\begin{proof}
The block structure in \eqref{eq:Gtilde} gives
\[
\boldsymbol\alpha^\T\widetilde{\boldsymbol G}\boldsymbol\alpha
={\Delta\over4}\Big(\sum_{j\in I_1\cup I_2}\alpha_j\Big)^2
+\tr(\boldsymbol\Sigma_1)\sum_{j\in I_1}\alpha_j^2
+\tr(\boldsymbol\Sigma_2)\sum_{j\in I_2}\alpha_j^2
+\sum_{k\in O}D_k\alpha_k^2.
\]
Lemma~\ref{lem:weighted-cs} minimizes the three diagonal sums, for fixed $(a,b,s)$, at the allocation \eqref{eq:allocation}; the lower bounds are $\tau_1a^2$, $\tau_2b^2$, and $\kappa s^2$. Since $s=a-b$, the result follows.
\end{proof}

Writing $\boldsymbol z=(a,b)^\T$,
\[
Q(a,b)=\boldsymbol z^\T\boldsymbol M\boldsymbol z,
\qquad
\boldsymbol M=
\begin{pmatrix}p+\tau_1+\kappa&p-\kappa\\p-\kappa&p+\tau_2+\kappa\end{pmatrix}.
\]
Its determinant is
\begin{equation}\label{eq:detM}
\det\boldsymbol M=\kappa\Delta_*+E,
\qquad
E={\Delta\over4}(\tau_1+\tau_2)+\tau_1\tau_2.
\end{equation}
Hence $\boldsymbol M$ is positive definite and $F(\boldsymbol z)=\boldsymbol g^\T\boldsymbol z-\boldsymbol z^\T\boldsymbol M\boldsymbol z/2$, with $\boldsymbol g=(2,0)^\T$, is strictly concave.

\subsection{The clean case}
\begin{proposition}[\citealp{nakayama2017}]\label{prop:clean}
Without outliers, or with $s=0$, the maximum equals $2/\Delta_*$ and is attained at $a=b=2/\Delta_*$, with $\alpha_j=2/(\Delta_*n_i)$ for $j\in I_i$.
\end{proposition}
\begin{proof}
With $b=a$, $Q(a,a)=\Delta_*a^2$ and
$F(a,a)=2a-\Delta_*a^2/2$; completing the square proves the claim.
\end{proof}

%% file: sections/05_one_sided.tex
\section{One-sided contamination: the dual problem}

The symbols $a^\circ$ and $b^\circ$ used below denote the coordinates of the unconstrained stationary point
\begin{equation}\label{eq:circ-def}
\binom{a^\circ}{b^\circ}:=\boldsymbol M^{-1}\boldsymbol g,
\end{equation}
whereas $(\astar,\bstar)$ denotes the constrained maximizer of $F$ over $\cD$. This distinction matters when $b^\circ<0$.

\begin{theorem}[Exact solution of the limiting dual]\label{thm:one-sided}
Let $E$ be defined in \eqref{eq:detM}.
\begin{enumerate}
\item[(i)] If $\kappa\ge\Delta/4$, the unique maximizer is the interior point $(\astar,\bstar)=(a^\circ,b^\circ)$, where
\begin{equation}\label{eq:one-int}
\astar={2(\Delta/4+\tau_2+\kappa)\over\kappa\Delta_*+E},\qquad
\bstar={2(\kappa-\Delta/4)\over\kappa\Delta_*+E},\qquad
\sstar=\astar-\bstar={2(\Delta/2+\tau_2)\over\kappa\Delta_*+E}.
\end{equation}
The maximum is
\begin{equation}\label{eq:F-one}
F_{\max}=\astar={2\over\Delta_*}+{2(\Delta/2+\tau_2)^2\over\Delta_*(\kappa\Delta_*+E)}.
\end{equation}
\item[(ii)] If $\kappa<\Delta/4$, the unique maximizer lies on the boundary:
\begin{equation}\label{eq:one-boundary}
\bstar=0,
\qquad
\astar=\sstar={8\over\Delta+4\tau_1+4\kappa},
\qquad
F_{\max}={8\over\Delta+4\tau_1+4\kappa}.
\end{equation}
\end{enumerate}
\end{theorem}
\begin{proof}
The unconstrained stationarity equation is $\boldsymbol M\boldsymbol z=\boldsymbol g$. Direct inversion gives
\[
\boldsymbol M^{-1}={1\over\kappa\Delta_*+E}
\begin{pmatrix}p+\tau_2+\kappa&\kappa-p\\\kappa-p&p+\tau_1+\kappa\end{pmatrix},
\]
which yields \eqref{eq:one-int}. The constraint $a\ge b$ is always strict, whereas $b^\circ\ge0$ exactly when $\kappa\ge p=\Delta/4$. Strict concavity then proves part (i), and $F(\boldsymbol z^\circ)=\boldsymbol g^\T\boldsymbol z^\circ/2=a^\circ$. Expanding $a^\circ-2/\Delta_*$ gives \eqref{eq:F-one}.

If $\kappa<p$, the unconstrained point is infeasible. On $b=0$,
$F(a,0)=2a-(p+\tau_1+\kappa)a^2/2$, whose maximizer is \eqref{eq:one-boundary}. At that point
$\nabla F=(0,-(p-\kappa)\astar)^\T$, so
$\nabla F^\T(\boldsymbol z-\boldsymbol z^\star)\le0$ for every $\boldsymbol z\in\cD$. Concavity proves global optimality and strict concavity proves uniqueness.
\end{proof}

At the optimum, both regimes satisfy
\begin{equation}\label{eq:identity}
p(\astar+\bstar)+\tau_1\astar+\kappa\sstar=2.
\end{equation}

\begin{corollary}[When are outliers harmless?]\label{cor:harmless}
Let $R=F_{\max}/(2/\Delta_*)$. Then $R>1$. In the interior regime,
\[
R=1+{(\Delta/2+\tau_2)^2\over\kappa\Delta_*+E}.
\]
Consequently, $R\to1$ if and only if $(\kappa\Delta_*+E)/(\Delta/2+\tau_2)^2\to\infty$; in particular, $\kappa/\Delta_*\to\infty$ is sufficient. For $m$ outliers at a common distance $D$, this sufficient condition is $D/(m\Delta_*)\to\infty$.
\end{corollary}
At the hard-margin optimum, $\lVert\widehat{\boldsymbol w}\rVert^2=2\max L$; therefore the outliers shrink the margin by the factor $R^{-1/2}$.

\begin{corollary}[Allocation, breakdown, and critical number]\label{cor:allocation}
If $\kappa\ge\Delta/4$, the outliers carry fraction
\[
{\sstar\over\astar}={\Delta/2+\tau_2\over\Delta/4+\tau_2+\kappa}
\]
of the class-2 mass, and outlier $k$ receives $\alpha_k=\sstar\kappa/D_k$. If $\kappa<\Delta/4$, then $\bstar=0$: every class-2 inlier has zero limiting weight. If $D_1=\cdots=D_m=D$, this collapse occurs exactly when $m>4D/\Delta$.
\end{corollary}

\begin{remark}
The threshold $\kappa=\Delta/4$ compares the effective outlier distance from the midpoint with the squared distance of the class-2 mean from that midpoint. Below the threshold, only outliers remain as class-2 support vectors. Because $D_k$ includes $\tr(\boldsymbol\Sigma_{0k})$, a noisy outlier can remain effectively distant even when its mean is at the midpoint.
\end{remark}

\begin{theorem}[Transfer of the optimal value]\label{thm:transfer-value}
Under (A1)--(A3),
\[
\max_{\boldsymbol\alpha\in\cC}L(\boldsymbol\alpha)
=F_{\max}\{1+\op(1)\}.
\]
\end{theorem}
\begin{proof}
Lemma~\ref{lem:geometry}, nonnegativity of $\boldsymbol\alpha$, and Lemma~\ref{lem:reduction} imply
$|\boldsymbol\alpha^\T\boldsymbol R\boldsymbol\alpha|
\le\epsilon\boldsymbol\alpha^\T\widetilde{\boldsymbol G}\boldsymbol\alpha$ uniformly on $\cC$, with $\epsilon=\op(1)$; (A3) supplies the uniform constant needed to compare the outlier-diagonal and inlier terms. On $\{\epsilon<1\}$,
\[
\one^\T\boldsymbol\alpha-\frac{1+\epsilon}{2}\boldsymbol\alpha^\T\widetilde{\boldsymbol G}\boldsymbol\alpha
\le L(\boldsymbol\alpha)\le
\one^\T\boldsymbol\alpha-\frac{1-\epsilon}{2}\boldsymbol\alpha^\T\widetilde{\boldsymbol G}\boldsymbol\alpha.
\]
Because $\cC$ is a cone, rescaling $\boldsymbol\alpha$ shows that the maxima of the two bounding objectives are $F_{\max}/(1+\epsilon)$ and $F_{\max}/(1-\epsilon)$. The result follows by squeezing.
\end{proof}

%% file: sections/06_discriminant.tex
\section{One-sided contamination: solution and discriminant}
Let $\widetilde{\boldsymbol\alpha}$ be the unique maximizer of $\widetilde L$ on $\cC$, obtained from \eqref{eq:allocation} at $(\astar,\bstar)$, and let $\widehat{\boldsymbol\alpha}$ maximize $L$. Define the seminorm
$\lVert\boldsymbol\theta\rVert_{\widetilde G}=(\boldsymbol\theta^\T\widetilde{\boldsymbol G}\boldsymbol\theta)^{1/2}$.

\begin{lemma}\label{lem:quad-facts}
For $c'>0$, let $f(\boldsymbol\alpha)=\one^\T\boldsymbol\alpha-(c'/2)\boldsymbol\alpha^\T\widetilde{\boldsymbol G}\boldsymbol\alpha$ and let $\boldsymbol\alpha^\circ$ maximize $f$ on $\cC$. Then
\[
f(\boldsymbol\alpha)\le f(\boldsymbol\alpha^\circ)-{c'\over2}\lVert\boldsymbol\alpha-\boldsymbol\alpha^\circ\rVert_{\widetilde G}^2.
\]
If $\boldsymbol A$ is positive semidefinite and $\boldsymbol\alpha^\circ$ maximizes $\one^\T\boldsymbol\alpha-\boldsymbol\alpha^\T\boldsymbol A\boldsymbol\alpha/2$ on the cone $\cC$, then
$\one^\T\boldsymbol\alpha^\circ=(\boldsymbol\alpha^\circ)^\T\boldsymbol A\boldsymbol\alpha^\circ$ and the maximum equals $\one^\T\boldsymbol\alpha^\circ/2$.
\end{lemma}
\begin{proof}
The first statement is the exact quadratic expansion around $\boldsymbol\alpha^\circ$ plus first-order optimality. For the second, maximize the objective along $t\boldsymbol\alpha^\circ$, $t\ge0$, and differentiate at $t=1$.
\end{proof}

\begin{theorem}[Convergence of the dual solution]\label{thm:dual-solution}
Assume (A1)--(A3) and put $\boldsymbol\delta=\widehat{\boldsymbol\alpha}-\widetilde{\boldsymbol\alpha}$. Then
\[
{\lVert\boldsymbol\delta\rVert_{\widetilde G}^2\over F_{\max}}=\op(1).
\]
If $\widehat a,\widehat b,\widehat s$ are the classwise sums of $\widehat{\boldsymbol\alpha}$, then
\[
\Delta|\widehat a-\astar+\widehat b-\bstar|=\op(1),\quad
\tau_1|\widehat a-\astar|=\op(1),\quad
\tau_2|\widehat b-\bstar|=\op(1),
\]
and, when $\kappa/\Delta_*$ is bounded, $\kappa|\widehat s-\sstar|=\op(1)$.
\end{theorem}
\begin{proof}
The sandwich used in Theorem~\ref{thm:transfer-value}, together with Lemma~\ref{lem:quad-facts}, gives an upper bound of order $\epsilon F_{\max}$ for
$\lVert\widehat{\boldsymbol\alpha}-\widetilde{\boldsymbol\alpha}\rVert_{\widetilde G}^2$; here $\epsilon=\op(1)$. Applying Lemma~\ref{lem:reduction} to the difference vector bounds each squared classwise discrepancy by this seminorm. The displayed rates follow from $\Delta_*F_{\max}=O(1)$ and the identity $\widehat s-\sstar=(\widehat a-\astar)-(\widehat b-\bstar)$.
\end{proof}

\subsection{The discriminant function}
For a new observation $\boldsymbol x_0$ from $\Pi_i$, independent of the training data, translation invariance gives
\begin{equation}\label{eq:disc-centered}
\widehat y(\boldsymbol x_0)=
\sum_{j=1}^{N}\widehat\alpha_jt_j(\boldsymbol x_j-\boldsymbol\mu_*)^\T(\boldsymbol x_0-\boldsymbol\mu_*)+\widehat c,
\qquad
\widehat c=\widehat b+\widehat{\boldsymbol w}^{\T}\boldsymbol\mu_*.
\end{equation}
For a training point, the KKT condition is
\begin{equation}\label{eq:kkt-centered}
(\boldsymbol G\widehat{\boldsymbol\alpha})_j+t_j\widehat c=1
\qquad\text{whenever }\widehat\alpha_j>0.
\end{equation}
Define
\begin{equation}\label{eq:cystar}
c^*=1-\kappa\sstar,
\qquad
y_i^*=(-1)^i{\Delta\over4}(\astar+\bstar)+c^*,\quad i=1,2.
\end{equation}

\begin{theorem}[Limit of the discriminant function]\label{thm:discriminant}
Under (A1)--(A3), $\widehat c=c^*+\op(1)$ and, for $\boldsymbol x_0\in\Pi_i$,
$\widehat y(\boldsymbol x_0)=y_i^*+\op(1)$. Explicitly, if $\kappa\ge\Delta/4$,
\begin{equation}\label{eq:y-interior}
y_i^*={(-1)^i\Delta(\tau_2+2\kappa)/2+\kappa(\tau_1-\tau_2)+E\over\kappa\Delta_*+E},
\end{equation}
and if $\kappa<\Delta/4$,
\begin{equation}\label{eq:y-boundary}
y_i^*={(-1)^i\Delta/2+\Delta/4+\tau_1-\kappa\over\Delta/4+\tau_1+\kappa}.
\end{equation}
\end{theorem}
\begin{proof}
Summing \eqref{eq:kkt-centered} over class-1 support vectors and dividing by their positive total gives
$\widehat c=\widehat a^{-1}\sum_{j\in I_1}\widehat\alpha_j(\boldsymbol G\widehat{\boldsymbol\alpha})_j-1$.
Lemma~\ref{lem:geometry}, Theorem~\ref{thm:dual-solution}, and the optimal allocation imply
$\widehat c=p(\astar+\bstar)+\tau_1\astar-1+\op(1)$.
Identity \eqref{eq:identity} turns this expression into $1-\kappa\sstar+\op(1)$.

For the data term in \eqref{eq:disc-centered}, the same moment bounds as in Lemma~\ref{lem:geometry} give
$\sum_j\widehat\alpha_jt_j(\boldsymbol x_j-\boldsymbol\mu_*)^\T(\boldsymbol x_0-\boldsymbol\mu_*)=(-1)^ip(\astar+\bstar)+\op(1)$.
Substitution proves the first claim. Equations \eqref{eq:y-interior} and \eqref{eq:y-boundary} follow by inserting Theorem~\ref{thm:one-sided} into \eqref{eq:cystar}.
\end{proof}

\begin{corollary}[Misclassification rates]\label{cor:misclass-one}
Let $e(i)$ be the probability that the SVM misclassifies an independent observation from $\Pi_i$. If $\limsup y_1^*<0$, then $e(1)\to0$, whereas if $\liminf y_1^*>0$, then $e(1)\to1$. Likewise, $e(2)\to0$ if $\liminf y_2^*>0$ and $e(2)\to1$ if $\limsup y_2^*<0$.

If $\tau_1=\tau_2=\tau$, define
\begin{equation}\label{eq:kappa-c}
\kappa_c=
\begin{cases}
\tau^2/\Delta,&\tau\ge\Delta/2,\\
\max(\tau-\Delta/4,0),&\tau<\Delta/2.
\end{cases}
\end{equation}
Then $y_2^*>0$ for every $\kappa$, while $y_1^*>0$ exactly when $\kappa<\kappa_c$. Thus, if $\limsup\kappa/\kappa_c<1$, the SVM misclassifies the clean class $\Pi_1$ with probability tending to one.
\end{corollary}

\begin{remark}
Although the outliers are labeled as class 2, they can destroy class 1 through the intercept. KKT places each active outlier on the positive margin. Because the weight vector is asymptotically orthogonal to the outlier direction, a small effective distance is accommodated mainly by raising the intercept and shifting the whole hyperplane toward $\Pi_1$.
\end{remark}

%% file: sections/07_two_sided.tex
\section{Two-sided contamination}\label{sec:two-sided}
We now allow both classes to contain outliers. Let $O_1$ and $O_2$ index outliers labeled as class 1 and class 2, with $m_i=|O_i|$ fixed. For every outlier, retain
$\boldsymbol\nu_k=\boldsymbol\mu_{0k}-\boldsymbol\mu_*$ and
$D_k=\lVert\boldsymbol\nu_k\rVert^2+\tr(\boldsymbol\Sigma_{0k})$, and define
\begin{equation}\label{eq:kappai}
\kappa_i=\left(\sum_{k\in O_i}D_k^{-1}\right)^{-1},
\qquad \kappa_i=\infty\text{ if }O_i=\varnothing.
\end{equation}
Assume (A2) for all outliers and add:
\begin{description}
\item[(A4)] For each $i$ with $O_i\ne\varnothing$,
$0<\liminf\kappa_i/\Delta_*\le\limsup\kappa_i/\Delta_*<\infty$.
\end{description}
The sign flip for a class-1 outlier changes only negligible off-diagonal terms, so Lemma~\ref{lem:geometry} remains valid.

\subsection{Reduction and KKT system}
Let $a,b,s_1,s_2$ be the total dual masses on $I_1,I_2,O_1,O_2$. The balance constraint is
$-a-s_1+b+s_2=0$. Put
$A=a+s_1=b+s_2$, so $\one^\T\boldsymbol\alpha=2A$.

\begin{lemma}[Reduction to four variables]\label{lem:four}
Let
\[
Q_2(a,b,s_1,s_2)=p(a+b)^2+\tau_1a^2+\tau_2b^2+\kappa_1s_1^2+\kappa_2s_2^2.
\]
Every feasible $\boldsymbol\alpha$ satisfies
$\boldsymbol\alpha^\T\widetilde{\boldsymbol G}\boldsymbol\alpha\ge Q_2$, with equality when mass is spread evenly within $I_i$ and as $\alpha_k=s_i\kappa_i/D_k$ within $O_i$. Consequently, the limiting maximum is the maximum of
$F_2=a+b+s_1+s_2-Q_2/2$ over
\[
\cD_2=\{a,b,s_1,s_2\ge0:a+s_1=b+s_2\}.
\]
\end{lemma}
The proof repeats Lemma~\ref{lem:reduction} group by group.

Attach multiplier $c$ to $a+s_1-b-s_2=0$ and put $P=p(a+b)$. The KKT system is
\begin{align}\label{eq:kkt-two}
P+\tau_1a&=1+c &&(a>0),& P+\tau_1a&\ge1+c&&(a=0),\nonumber\\
P+\tau_2b&=1-c &&(b>0),& P+\tau_2b&\ge1-c&&(b=0),\\
\kappa_1s_1&=1+c,&&& \kappa_2s_2&=1-c.&&\nonumber
\end{align}
At the unique maximizer, $-1<c<1$, both $s_i$ are positive, and $a,b$ cannot both vanish.

Write
\begin{equation}\label{eq:two-shorthands}
H=\kappa_1\kappa_2\Delta_*+(\kappa_1+\kappa_2)E,
\quad
\rho_i=\left({1\over p+\tau_i}+{1\over\kappa_i}\right)^{-1},
\quad
\beta_i(x)={px\over x+p+\tau_{3-i}}.
\end{equation}

\begin{theorem}[Exact solution under two-sided contamination]\label{thm:two-sided}
Let $(\astar,\bstar,s_1^*,s_2^*)$ maximize $F_2$, let $c^*$ be the multiplier, and let $P^*=p(\astar+\bstar)$.
\begin{description}
\item[(N)] If $\kappa_1\ge\beta_1(\kappa_2)$ and $\kappa_2\ge\beta_2(\kappa_1)$, then
\begin{align}
 c^*&={\kappa_1\kappa_2(\tau_1-\tau_2)+(\kappa_1-\kappa_2)E\over H},
& P^*&={ (\Delta/2)(2\kappa_1\kappa_2+\kappa_1\tau_2+\kappa_2\tau_1)\over H},\nonumber\\
 \astar&={1+c^*-P^*\over\tau_1},&
 \bstar&={1-c^*-P^*\over\tau_2},\nonumber\\
 s_1^*&={1+c^*\over\kappa_1},&
 s_2^*&={1-c^*\over\kappa_2}.\label{eq:Nsol}
\end{align}
with
\[
\max F_2={2\{(\kappa_1+p+\tau_1)(\kappa_2+p+\tau_2)-p^2\}\over H}.
\]
\item[(B2)] If $\kappa_2<\beta_2(\kappa_1)$, then $\bstar=0$ and
\[
c^*={\rho_1-\kappa_2\over\rho_1+\kappa_2},\quad
\astar={1+c^*\over p+\tau_1},\quad
s_1^*={1+c^*\over\kappa_1},\quad
s_2^*={1-c^*\over\kappa_2},\quad
\max F_2={2\over\rho_1+\kappa_2}.
\]
\item[(B1)] If $\kappa_1<\beta_1(\kappa_2)$, exchange the class labels in (B2): $\astar=0$, $c^*=-(\rho_2-\kappa_1)/(\rho_2+\kappa_1)$, $\bstar=(1-c^*)/(p+\tau_2)$, and $\max F_2=2/(\rho_2+\kappa_1)$.
\end{description}
The B1 and B2 conditions are mutually exclusive.
\end{theorem}
\begin{proof}
In regime N, \eqref{eq:kkt-two} gives
$a=(1+c-P)/\tau_1$, $b=(1-c-P)/\tau_2$, $s_1=(1+c)/\kappa_1$, and $s_2=(1-c)/\kappa_2$. Substitution into $P=p(a+b)$ and the balance equation yields the displayed linear solution for $(c,P)$ and hence \eqref{eq:Nsol}. Lemma~\ref{lem:quad-facts} gives the maximum from half the total mass. The nonnegativity conditions reduce to the two stated inequalities.

In B2, set $b=0$. Then $P=pa$, $(p+\tau_1)a=1+c$, and balance gives $(1+c)/\rho_1=(1-c)/\kappa_2$. This yields the formulas; the inactive KKT inequality is precisely $\kappa_2<\beta_2(\kappa_1)$. B1 is symmetric. Finally, $\beta_i(x)<x$, so both strict breakdown inequalities cannot hold simultaneously.
\end{proof}

\begin{corollary}[Breakdown under two-sided contamination]\label{cor:breakdown-two}
Class 2 collapses if and only if $\kappa_2<\beta_2(\kappa_1)=(\Delta/4)\kappa_1/(\kappa_1+\Delta/4+\tau_1)$, and symmetrically for class 1. Hence at most one class collapses. The boundary $\beta_2(\kappa_1)$ increases in $\kappa_1$, so class-1 outliers make class 2 harder to collapse. Under symmetric contamination, neither class collapses.
\end{corollary}

\subsection{Transfer and classification}
\begin{theorem}\label{thm:transfer-two}
Assume (A1)--(A4). Then the actual optimum is $\max F_2\{1+\op(1)\}$, its four classwise sums converge at the corresponding weighted rates, $\widehat c=c^*+\op(1)$, and for $\boldsymbol x_0\in\Pi_i$,
\begin{equation}\label{eq:y-two}
\widehat y(\boldsymbol x_0)=y_i^*+\op(1),
\qquad y_i^*=(-1)^iP^*+c^*.
\end{equation}
\end{theorem}
\begin{proof}
The sandwich and strong-concavity arguments of Theorems~\ref{thm:transfer-value} and \ref{thm:dual-solution} apply after replacing Lemma~\ref{lem:reduction} by Lemma~\ref{lem:four}. Assumption (A4) keeps the relevant costs comparable with $\Delta_*$. Summing the actual KKT equations over either outlier group gives $1+t_i\widehat c=\kappa_i\widehat s_i+\op(1)$, and the limit is \eqref{eq:kkt-two}. The data term is unchanged because outlier--test cross terms remain negligible.
\end{proof}

\begin{corollary}[Misclassification under two-sided contamination]\label{cor:misclass-two}
Assume $\tau_1=\tau_2=\tau$. For $x>0$, define
\begin{equation}\label{eq:kappacx}
\kappa_c(x)=
\begin{cases}
\displaystyle {\tau^2x\over\tau^2+\Delta x+\Delta\tau},&\tau\ge\Delta/2,\\[2mm]
\displaystyle \max\left\{{(\tau-\Delta/4)x\over x+\Delta/4+\tau},0\right\},&\tau<\Delta/2.
\end{cases}
\end{equation}
Then $y_1^*>0$ exactly when $\kappa_2<\kappa_c(\kappa_1)$ and $y_2^*<0$ exactly when $\kappa_1<\kappa_c(\kappa_2)$. Thus at most one class is asymptotically misclassified. Moreover, $\kappa_c(x)$ increases in $x$ and approaches the one-sided threshold \eqref{eq:kappa-c} as $x\to\infty$.
\end{corollary}

\begin{corollary}[Symmetric contamination]\label{cor:symmetric}
If $\tau_1=\tau_2=\tau$ and $\kappa_1=\kappa_2=\kappa$, then
\[
c^*=0,\qquad \astar=\bstar={2\over\Delta_*},\qquad
s_1^*=s_2^*={1\over\kappa},\qquad
\max F_2={2\over\Delta_*}+{1\over\kappa},\qquad
 y_i^*=(-1)^i{\Delta\over\Delta_*}.
\]
Thus the limiting discriminant equals that of the clean SVM, although the dual optimum increases and the margin shrinks.
\end{corollary}

\begin{remark}
Each active outlier group imposes a margin equation, $\kappa_1s_1=1+c$ and $\kappa_2s_2=1-c$. The two equations pull the intercept in opposite directions. Equal pulls cancel; unequal pulls shift failure toward the class with the more distant outliers.
\end{remark}

%% file: sections/08_simulations.tex
\section{Numerical experiments}\label{sec:simulations}
All experiments use $n_1=n_2=10$ and $\boldsymbol\Sigma_1=\boldsymbol\Sigma_2=\boldsymbol I_d$, so $\tau_1=\tau_2=d/10$. The first $d/2$ entries of $\boldsymbol\mu$ are equal and the remaining entries are zero, with $\lVert\boldsymbol\mu\rVert^2=\Delta$. Each outlier has covariance $\sigma_0^2\boldsymbol I_d$ and an offset supported on a separate block of the last $d/2$ coordinates, so (A2) holds. The offset norm is chosen to attain the target effective distance.

The computation solves the hard-margin dual directly. Numerically, it optimizes $\boldsymbol\gamma=d\boldsymbol\alpha$ against the normalized signed Gram matrix; this is equivalent to using a soft-margin upper bound so large that it is never active. The complete implementation, fixed seed, raw CSV files, software versions, and figure-generation code are included in \texttt{experiments/run\_experiments.py} and \texttt{results/}.

\paragraph{Why the replication counts are adequate here.}
The experiments are validation checks of deterministic asymptotic formulas, not Monte Carlo estimates of an unknown scientific effect. Each experiment compares an entire parameter grid against a prespecified curve or phase boundary, and the predicted changes are large (including exactly zero limiting inlier mass and near-unit changes in error probability). Ten replications are used for Experiments 1 and 4, twenty for the convergence-rate study, eight for Experiments 5 and 6, and six for Experiments 3 and 7. These counts are intentionally modest because one replication contains up to forty thousand coordinates and the aim is to check structural agreement rather than to report high-precision simulation estimates. Reproducibility is provided by a fixed seed, and Experiment 2 additionally reports Monte Carlo standard errors in its CSV output. The conclusions are based on agreement across grid points and across seven distinct checks, not on significance testing at a single point.

\subsection{Experiment 1: dual value and weight allocation}
\begin{figure}[htbp]
\centering
\includegraphics[width=.72\textwidth]{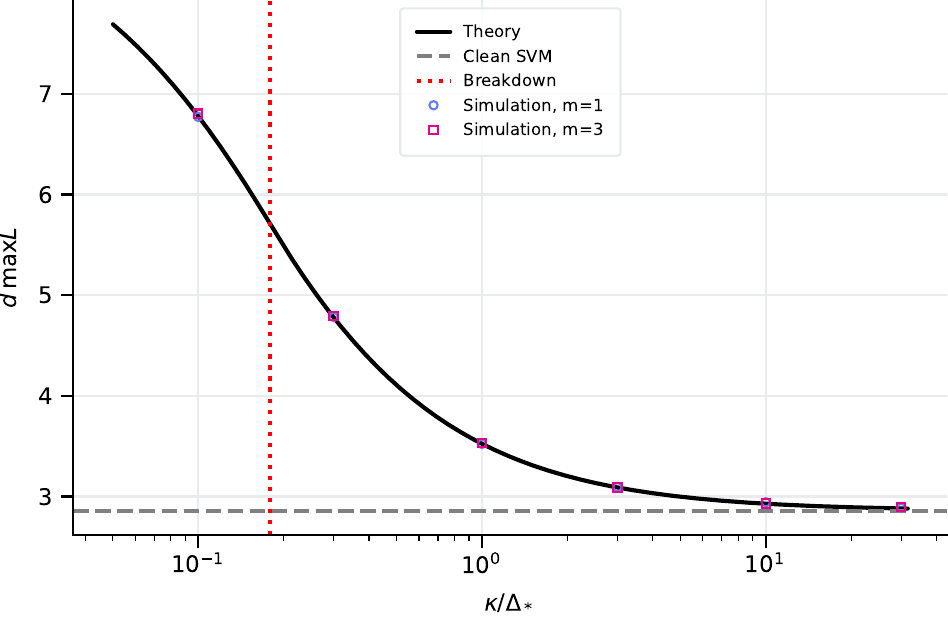}
\caption{Experiment 1: dual optimum against the effective-distance ratio. The vertical dotted line is the breakdown threshold.}
\label{fig:exp1}
\end{figure}

Set $d=20{,}000$, $\Delta=10{,}000$, and hence $\Delta_*=14{,}000$. Add $m\in\{1,3\}$ outliers with $\sigma_0=0.2$ and choose their common distance so that $\kappa=r\Delta_*$ for $r\in\{0.1,0.3,1,3,10,30\}$. Breakdown occurs at $\kappa/\Delta_*\approx0.179$. Table~\ref{tab:exp1} and Figure~\ref{fig:exp1} compare ten-replication means with Theorem~\ref{thm:one-sided}. The two values of $m$ produce essentially the same results, as predicted by the reduction to $\kappa$.
\begin{table}[htbp]
\centering\small
\caption{Experiment 1: simulated and theoretical dual quantities, multiplied by $d$.}
\label{tab:exp1}
\resizebox{\textwidth}{!}{\input{results/experiment1_table.tex}}
\end{table}

\subsection{Experiment 2: convergence in dimension}
\begin{figure}[htbp]
\centering
\includegraphics[width=.72\textwidth]{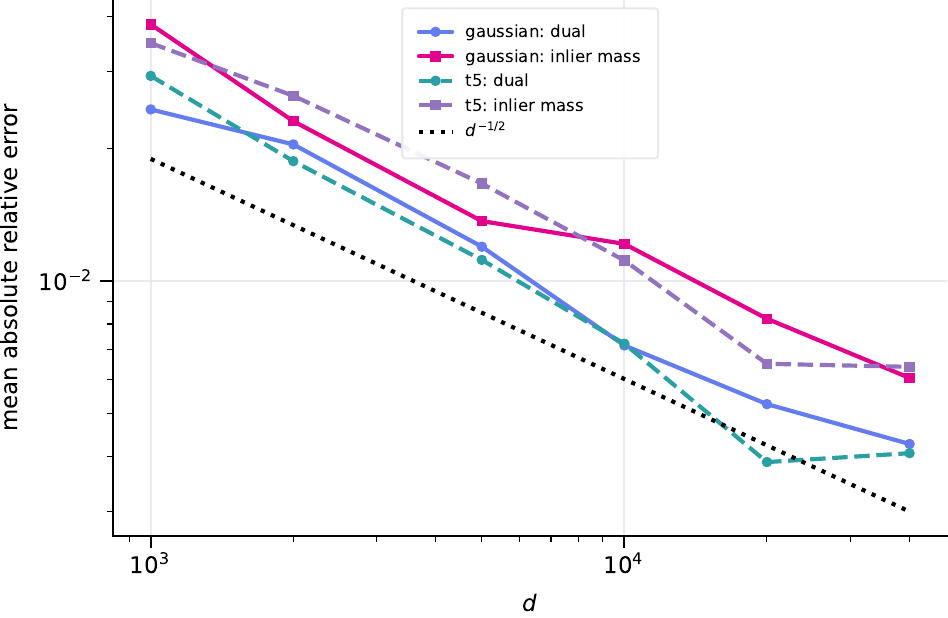}
\caption{Experiment 2: mean absolute relative errors against $d$ on log--log axes.}
\label{fig:exp2}
\end{figure}

Vary $d$ from $1{,}000$ to $40{,}000$, set $\Delta=d/2$, $m=1$, and $\kappa=\Delta_*$. Over twenty replications, record the mean absolute relative errors of the dual value and class-2 inlier mass. Alongside Gaussian noise, use a standardized $t_5$ distribution. Figure~\ref{fig:exp2} shows decay under both distributions at the order suggested by the Chebyshev bounds in Lemma~\ref{lem:geometry}.

\subsection{Experiment 3: critical number of outliers}
Set $d=20{,}000$, $\Delta=10{,}000$, and give every outlier common effective distance $D=2\Delta$. Then $\kappa=D/m$, and Corollary~\ref{cor:allocation} predicts collapse for $m>8$. Table~\ref{tab:exp3}, based on six replications, shows the class-2 inlier mass approaching zero at the predicted boundary.
\begin{table}[htbp]
\centering\small
\caption{Experiment 3: dependence on the number of outliers at fixed individual effective distance.}
\label{tab:exp3}
\resizebox{\textwidth}{!}{\input{results/experiment3_table.tex}}
\end{table}

\subsection{Experiment 4: discriminant function and misclassification}
\begin{figure}[htbp]
\centering
\includegraphics[width=.92\textwidth]{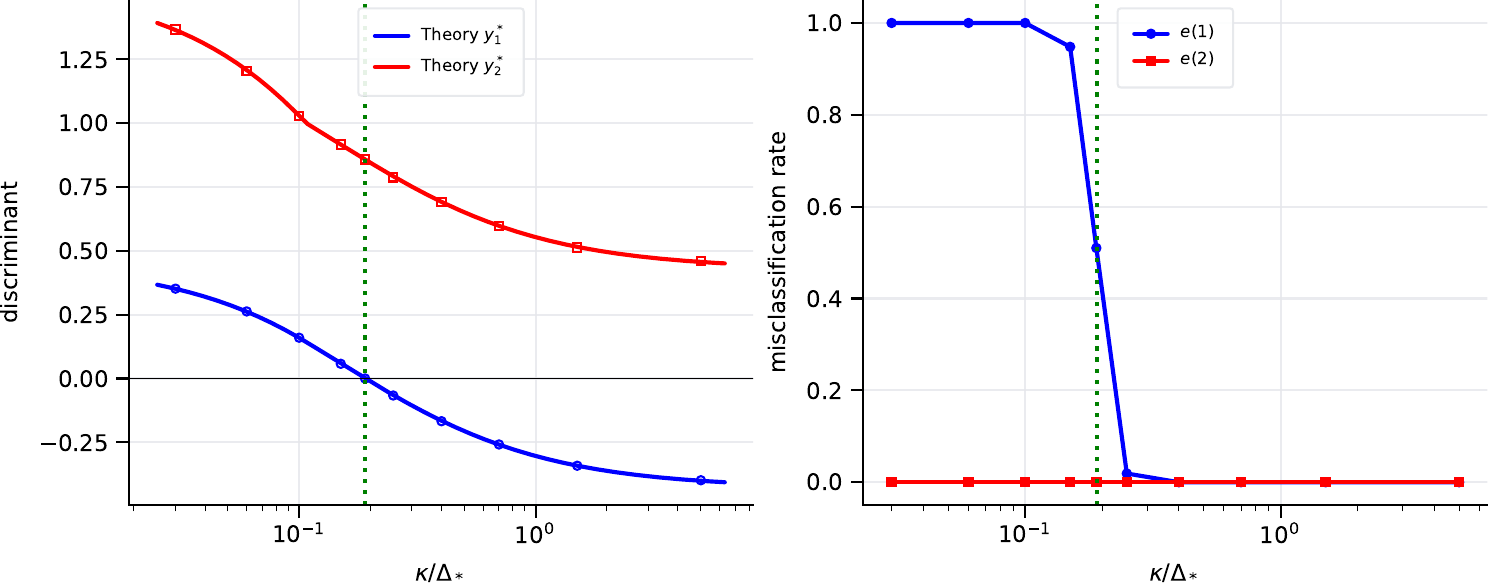}
\caption{Experiment 4. Left: limiting and simulated discriminant values. Right: misclassification rates. The vertical dotted line is the theoretical transition.}
\label{fig:exp4}
\end{figure}

Set $d=20{,}000$, $\Delta=3{,}000$, and $\tau=2{,}000$, so $\Delta_*=7{,}000$ and $\kappa_c/\Delta_*\approx0.19$. Add one outlier with $\sigma_0=0.05$, vary $\kappa/\Delta_*$ from $0.03$ to $5$, and evaluate 100 test scores per class in each of ten replications. Table~\ref{tab:exp4} and Figure~\ref{fig:exp4} show the predicted loss of class 1 near $\kappa_c$, while class 2 remains correctly classified.
\begin{table}[htbp]
\centering\small
\caption{Experiment 4: test-score means and misclassification rates.}
\label{tab:exp4}
\resizebox{\textwidth}{!}{\input{results/experiment4_table.tex}}
\end{table}

\subsection{Experiment 5: one-sided versus symmetric contamination}
\begin{figure}[htbp]
\centering
\includegraphics[width=.92\textwidth]{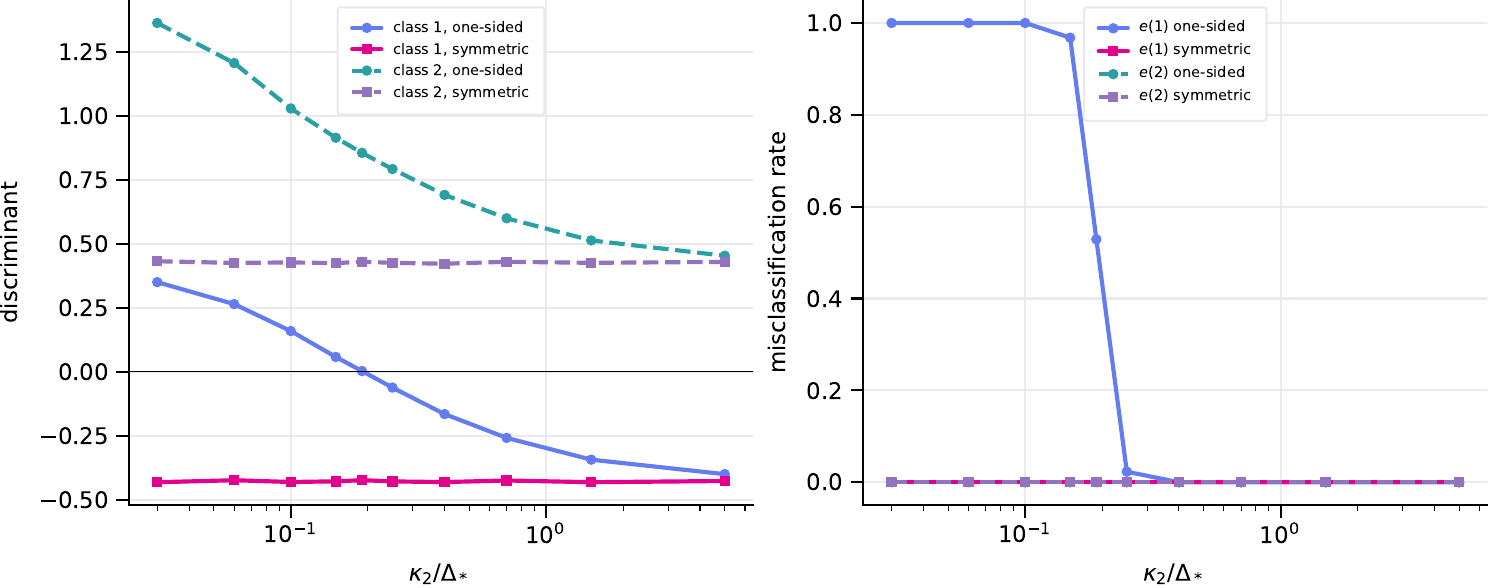}
\caption{Experiment 5. Left: discriminant values. Right: misclassification rates. Separate axes and reserved caption spacing prevent graphical elements from touching the caption.}
\label{fig:exp5}
\end{figure}

Repeat Experiment 4 with eight replications, comparing the one-sided design with a second, mirror outlier in class 1 having the same effective distance. Under symmetric contamination, the discriminant remains at the clean limit even when the outliers carry most of the dual mass; Figure~\ref{fig:exp5} illustrates the cancellation in Corollary~\ref{cor:symmetric}.
\begin{table}[htbp]
\centering\scriptsize
\caption{Experiment 5: one-sided and symmetric two-sided contamination.}
\label{tab:exp5}
\resizebox{\textwidth}{!}{\input{results/experiment5_combined_table.tex}}
\end{table}

\subsection{Experiment 6: asymmetric two-sided contamination}
\begin{figure}[htbp]
\centering
\includegraphics[width=.92\textwidth]{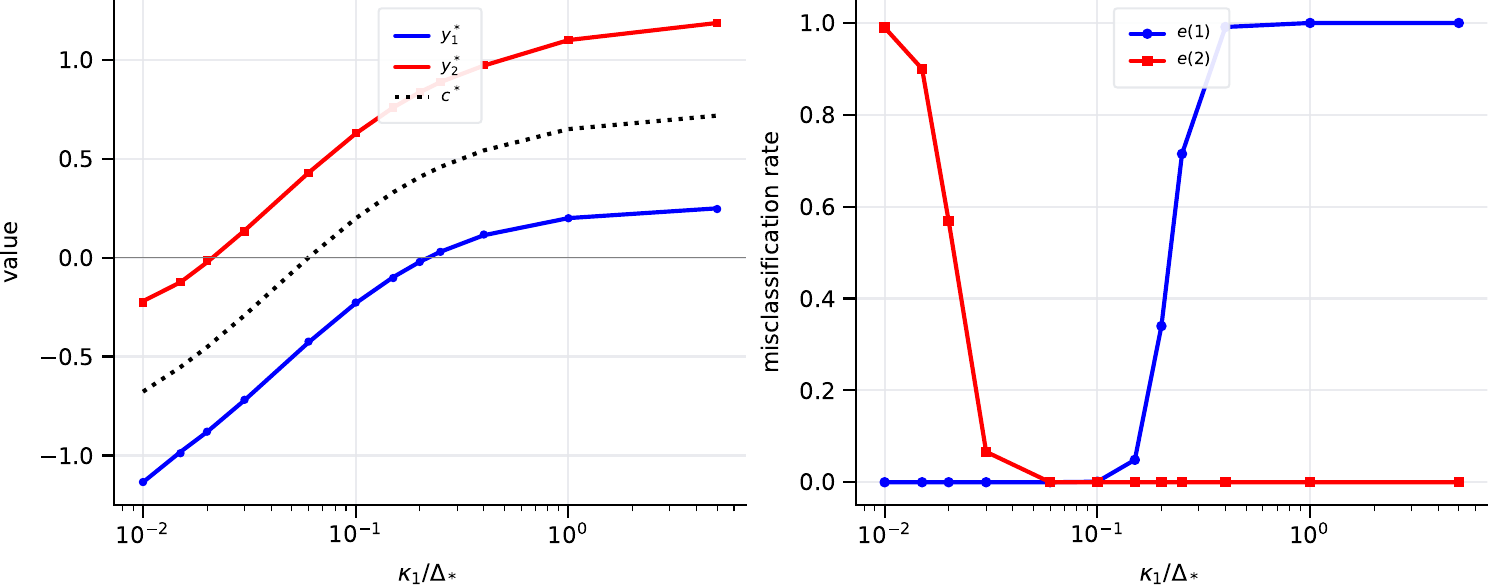}
\caption{Experiment 6. Left: discriminants and limiting intercept. Right: misclassification rates.}
\label{fig:exp6}
\end{figure}

\begin{figure}[htbp]
\centering
\includegraphics[width=.68\textwidth]{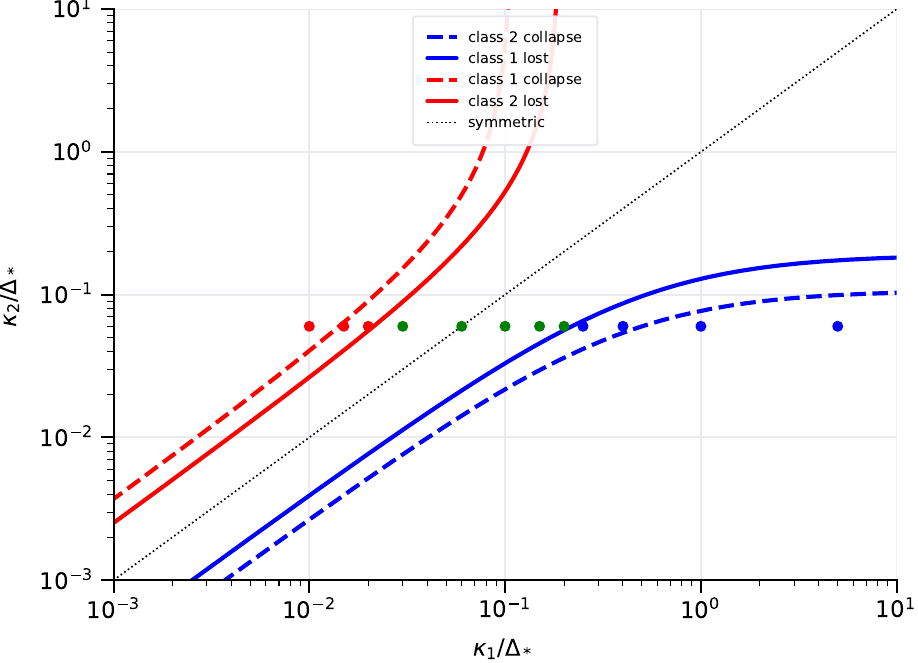}
\caption{Phase diagram in the $(\kappa_1,\kappa_2)$ plane. Solid curves are misclassification boundaries, dashed curves are support-collapse boundaries, and colored points are Experiment 6 outcomes.}
\label{fig:phase}
\end{figure}

Fix the class-2 outlier at $\kappa_2/\Delta_*=0.06$ and vary $\kappa_1/\Delta_*$. Corollary~\ref{cor:misclass-two} predicts three regions: loss of class 2 for very small $\kappa_1$, a central region in which both classes are correct, and loss of class 1 for larger $\kappa_1$. Table~\ref{tab:exp6}, Figure~\ref{fig:exp6}, and the phase diagram in Figure~\ref{fig:phase} confirm this movement of the intercept.
\begin{table}[htbp]
\centering\scriptsize
\caption{Experiment 6: asymmetric two-sided contamination.}
\label{tab:exp6}
\resizebox{\textwidth}{!}{\input{results/experiment6_table.tex}}
\end{table}

\subsection{Experiment 7: dual allocation in all three regimes}
Finally, set $d=20{,}000$, $\Delta=10{,}000$, and $\sigma_0=0.1$, and select ten pairs $(\kappa_1,\kappa_2)$ covering N, B1, and B2. Six-replication averages in Table~\ref{tab:exp7} agree with Theorem~\ref{thm:two-sided} and select the regime predicted by Corollary~\ref{cor:breakdown-two}.
\begin{table}[p]
\centering\scriptsize
\caption{Experiment 7: simulated and theoretical dual allocations, multiplied by $d$.}
\label{tab:exp7}
\resizebox{\textwidth}{!}{\input{results/experiment7_table.tex}}
\end{table}

%% file: results/experiment1_table.tex
\begin{tabular}{rrrrrrrr}
\toprule
$m$ & $\kappa/\Delta_\ast$ & $dL$ (sim.) & $dF$ (th.) & $db$ (sim.) & $db^\ast$ (th.) & $ds$ (sim.) & $ds^\ast$ (th.) \\
\midrule
1 & 0.100 & 6.771 & 6.780 & 0.000 & 0.000 & 6.771 & 6.780 \\
1 & 0.300 & 4.784 & 4.780 & 0.947 & 0.934 & 3.836 & 3.846 \\
1 & 1.000 & 3.522 & 3.524 & 2.193 & 2.190 & 1.329 & 1.333 \\
1 & 3.000 & 3.093 & 3.090 & 2.628 & 2.625 & 0.465 & 0.465 \\
1 & 10.000 & 2.942 & 2.928 & 2.800 & 2.786 & 0.142 & 0.142 \\
1 & 30.000 & 2.885 & 2.881 & 2.839 & 2.833 & 0.047 & 0.048 \\
3 & 0.100 & 6.802 & 6.780 & 0.000 & 0.000 & 6.802 & 6.780 \\
3 & 0.300 & 4.793 & 4.780 & 0.927 & 0.934 & 3.866 & 3.846 \\
3 & 1.000 & 3.532 & 3.524 & 2.200 & 2.190 & 1.332 & 1.333 \\
3 & 3.000 & 3.090 & 3.090 & 2.624 & 2.625 & 0.466 & 0.465 \\
3 & 10.000 & 2.930 & 2.928 & 2.789 & 2.786 & 0.141 & 0.142 \\
3 & 30.000 & 2.893 & 2.881 & 2.845 & 2.833 & 0.048 & 0.048 \\
\bottomrule
\end{tabular}

%% file: results/experiment3_table.tex
\begin{tabular}{rrrrrrrr}
\toprule
$m$ & $\kappa$ & $dL$ (sim.) & $dF$ (th.) & $db$ (sim.) & $db^\ast$ (th.) & $ds$ (sim.) & $ds^\ast$ (th.) \\
\midrule
1 & 20000.000 & 3.322 & 3.333 & 2.372 & 2.381 & 0.950 & 0.952 \\
2 & 10000.000 & 3.778 & 3.766 & 1.956 & 1.948 & 1.822 & 1.818 \\
4 & 5000.000 & 4.534 & 4.524 & 1.189 & 1.190 & 3.346 & 3.333 \\
6 & 3333.333 & 5.168 & 5.165 & 0.542 & 0.549 & 4.626 & 4.615 \\
8 & 2500.000 & 5.723 & 5.714 & 0.042 & 0.000 & 5.681 & 5.714 \\
10 & 2000.000 & 6.160 & 6.154 & 0.000 & 0.000 & 6.160 & 6.154 \\
12 & 1666.667 & 6.496 & 6.486 & 0.000 & 0.000 & 6.496 & 6.486 \\
16 & 1250.000 & 6.923 & 6.957 & 0.000 & 0.000 & 6.923 & 6.957 \\
\bottomrule
\end{tabular}

%% file: results/experiment4_table.tex
\begin{tabular}{rrrrrrr}
\toprule
$\kappa/\Delta_\ast$ & $\bar y_1$ (sim.) & $y_1^\ast$ (th.) & $\bar y_2$ (sim.) & $y_2^\ast$ (th.) & $e(1)$ & $e(2)$ \\
\midrule
0.030 & 0.352 & 0.351 & 1.367 & 1.365 & 1.000 & 0.000 \\
0.060 & 0.263 & 0.262 & 1.204 & 1.208 & 1.000 & 0.000 \\
0.100 & 0.160 & 0.159 & 1.029 & 1.029 & 1.000 & 0.000 \\
0.150 & 0.058 & 0.059 & 0.917 & 0.916 & 0.948 & 0.000 \\
0.190 & 0.000 & 0.001 & 0.858 & 0.858 & 0.510 & 0.000 \\
0.250 & -0.067 & -0.065 & 0.788 & 0.792 & 0.019 & 0.000 \\
0.400 & -0.167 & -0.165 & 0.691 & 0.692 & 0.000 & 0.000 \\
0.700 & -0.258 & -0.259 & 0.598 & 0.598 & 0.000 & 0.000 \\
1.500 & -0.342 & -0.342 & 0.513 & 0.516 & 0.000 & 0.000 \\
5.000 & -0.399 & -0.401 & 0.460 & 0.456 & 0.000 & 0.000 \\
\bottomrule
\end{tabular}

%% file: results/experiment5_combined_table.tex
\begin{tabular}{rrrrrrrrr}
\toprule
$\kappa/\Delta_\ast$ & $\bar y_1$ one & $y_1^\ast$ one & $e_1$ one & $e_2$ one & $\bar y_1$ sym. & $y_1^\ast$ sym. & $e_1$ sym. & $e_2$ sym. \\
\midrule
0.030 & 0.350 & 0.351 & 1.000 & 0.000 & -0.432 & -0.429 & 0.000 & 0.000 \\
0.060 & 0.265 & 0.262 & 1.000 & 0.000 & -0.424 & -0.429 & 0.000 & 0.000 \\
0.100 & 0.159 & 0.159 & 1.000 & 0.000 & -0.431 & -0.429 & 0.000 & 0.000 \\
0.150 & 0.058 & 0.059 & 0.968 & 0.000 & -0.429 & -0.429 & 0.000 & 0.000 \\
0.190 & 0.003 & 0.001 & 0.529 & 0.000 & -0.424 & -0.429 & 0.000 & 0.000 \\
0.250 & -0.062 & -0.065 & 0.023 & 0.000 & -0.429 & -0.429 & 0.000 & 0.000 \\
0.400 & -0.166 & -0.165 & 0.000 & 0.000 & -0.432 & -0.429 & 0.000 & 0.000 \\
0.700 & -0.259 & -0.259 & 0.000 & 0.000 & -0.425 & -0.429 & 0.000 & 0.000 \\
1.500 & -0.344 & -0.342 & 0.000 & 0.000 & -0.432 & -0.429 & 0.000 & 0.000 \\
5.000 & -0.401 & -0.401 & 0.000 & 0.000 & -0.427 & -0.429 & 0.000 & 0.000 \\
\bottomrule
\end{tabular}

%% file: results/experiment6_table.tex
\begin{tabular}{rrrrrrrr}
\toprule
$\kappa_1/\Delta_\ast$ & $\bar y_1$ (sim.) & $y_1^\ast$ (th.) & $\bar y_2$ (sim.) & $y_2^\ast$ (th.) & $c^\ast$ & $e(1)$ & $e(2)$ \\
\midrule
0.010 & -1.134 & -1.135 & -0.227 & -0.220 & -0.678 & 0.000 & 0.990 \\
0.015 & -0.989 & -0.982 & -0.122 & -0.125 & -0.554 & 0.000 & 0.900 \\
0.020 & -0.880 & -0.881 & -0.016 & -0.024 & -0.452 & 0.000 & 0.569 \\
0.030 & -0.719 & -0.721 & 0.132 & 0.136 & -0.292 & 0.000 & 0.066 \\
0.060 & -0.425 & -0.429 & 0.426 & 0.429 & 0.000 & 0.000 & 0.000 \\
0.100 & -0.227 & -0.231 & 0.631 & 0.627 & 0.198 & 0.001 & 0.000 \\
0.150 & -0.103 & -0.099 & 0.758 & 0.758 & 0.330 & 0.049 & 0.000 \\
0.200 & -0.022 & -0.022 & 0.836 & 0.836 & 0.407 & 0.340 & 0.000 \\
0.250 & 0.029 & 0.029 & 0.888 & 0.886 & 0.458 & 0.715 & 0.000 \\
0.400 & 0.116 & 0.113 & 0.973 & 0.970 & 0.541 & 0.991 & 0.000 \\
1.000 & 0.199 & 0.199 & 1.098 & 1.099 & 0.649 & 1.000 & 0.000 \\
5.000 & 0.245 & 0.249 & 1.184 & 1.185 & 0.717 & 1.000 & 0.000 \\
\bottomrule
\end{tabular}

%% file: results/experiment7_table.tex
\begin{tabular}{rrrrrrrrrrrrl}
\toprule
$\kappa_1/\Delta_\ast$ & $\kappa_2/\Delta_\ast$ & $dL$ (sim.) & $dF$ (th.) & $da$ (sim.) & $da^\ast$ (th.) & $db$ (sim.) & $db^\ast$ (th.) & $ds_1$ (sim.) & $ds_1^\ast$ (th.) & $ds_2$ (sim.) & $ds_2^\ast$ (th.) & regime \\
\midrule
1.000 & 1.000 & 4.297 & 4.286 & 2.862 & 2.857 & 2.872 & 2.857 & 1.435 & 1.429 & 1.426 & 1.429 & N \\
0.300 & 0.300 & 7.653 & 7.619 & 2.902 & 2.857 & 2.856 & 2.857 & 4.751 & 4.762 & 4.797 & 4.762 & N \\
0.100 & 0.100 & 17.147 & 17.143 & 2.833 & 2.857 & 2.862 & 2.857 & 14.314 & 14.286 & 14.285 & 14.286 & N \\
3.000 & 0.100 & 7.340 & 7.320 & 6.635 & 6.612 & 0.000 & 0.000 & 0.705 & 0.708 & 7.340 & 7.320 & B2 \\
1.000 & 0.100 & 8.333 & 8.324 & 6.314 & 6.299 & 0.000 & 0.000 & 2.019 & 2.025 & 8.333 & 8.324 & B2 \\
0.300 & 0.100 & 11.206 & 11.220 & 5.272 & 5.296 & 0.450 & 0.418 & 5.933 & 5.923 & 10.756 & 10.801 & N \\
0.100 & 0.300 & 11.213 & 11.220 & 0.440 & 0.418 & 5.286 & 5.296 & 10.773 & 10.801 & 5.927 & 5.923 & N \\
0.100 & 1.000 & 8.316 & 8.324 & 0.000 & 0.000 & 6.294 & 6.299 & 8.316 & 8.324 & 2.022 & 2.025 & B1 \\
0.100 & 3.000 & 7.311 & 7.320 & 0.000 & 0.000 & 6.597 & 6.612 & 7.311 & 7.320 & 0.714 & 0.708 & B1 \\
0.050 & 0.050 & 31.587 & 31.429 & 2.857 & 2.857 & 2.871 & 2.857 & 28.731 & 28.571 & 28.716 & 28.571 & N \\
\bottomrule
\end{tabular}

%% file: sections/09_discussion.tex
\section{Discussion}
In HDLSS settings, one effective outlier distance per class summarizes the leading effect of contamination on the hard-margin linear SVM. Under one-sided contamination, distant outliers leave the dual value, weights, and discriminant asymptotically unchanged. Closer outliers cause two distinct failures: below the support-collapse boundary, the contaminated class is represented only by its outliers; below the classification boundary, the intercept moves far enough to misclassify the clean class. Several moderately distant outliers can therefore have the same effect as one close outlier.

Two-sided contamination behaves differently. Outliers on opposite sides pull the limiting intercept in opposite directions. At most one class can collapse and at most one can be misclassified. The boundaries depend on the effective distance in the other class, and symmetric contamination leaves the limiting discriminant at its clean value. The dual optimum still increases and the margin shrinks, so margin diagnostics can reveal contamination even when classification is asymptotically unaffected.

Two practical implications follow. First, the diagonal of the centered Gram matrix estimates each observation's effective distance: it is asymptotically $D_k$ for an outlier and $\Delta/4+\tr(\boldsymbol\Sigma_i)$ for an inlier. Classwise effective distances therefore predict the sign and approximate size of the intercept shift through Theorem~\ref{thm:two-sided}. Second, because much of the damage is mediated by the intercept, a bias correction in the spirit of \citet{nakayama2017}, or a finite soft-margin bound that caps each observation's dual weight, should limit the effect.

Several questions remain. Assumption (A2) requires the outlier offsets to be asymptotically orthogonal to the population contrast. If their inner products are of the same order as $\Delta_*$, cross terms survive and the low-dimensional reductions must be modified. The lower bound in (A4) excludes outliers that lie exactly at the midpoint on the $\Delta_*$ scale; such points dominate the dual and require a different normalization. Extensions to kernel SVMs \citep{nakayama2020}, robust or bias-corrected variants, and comparisons with distance-weighted discrimination \citep{marron2007} are natural directions.